\documentclass{article}

\usepackage{microtype}
\usepackage{graphicx}
\usepackage{subfigure}
\usepackage{booktabs} 

\newcommand{\alglinelabel}{%
  \addtocounter{ALC@line}{-1}
  \refstepcounter{ALC@line}
  \label
}

\usepackage{amsmath}  
\usepackage{amssymb}  
\usepackage{mathtools}
\usepackage{amsthm} 
\usepackage{multirow}

\theoremstyle{plain}

\theoremstyle{definition}

\theoremstyle{remark}

\usepackage{mathtools}

\usepackage{nicefrac}

\usepackage{adjustbox}             
\usepackage{xcolor}

\newcommand{\fedht}{\algname{FedHT}}
\newcommand{\fediht}{\algname{FedIHT}}
\newcommand{\randprox}{\algname{RandProx}}
\newcommand{\proxskip}{\algname{ProxSkip}}
\newcommand{\randproxlone}{\algname{RandProx-l$_1$}}
\newcommand{\scaffold}{\algname{Scaffold}}
\newcommand{\fedavg}{\algname{FedAvg}}
\newcommand{\fedsparsify}{\algname{FedSparsify}}
\newcommand{\finaltopk}{\algname{Final-TopK}}
\newcommand{\ourmethod}{\algname{Sparse-ProxSkip}}
\newcommand{\ourlocal}{\algname{Sparse-ProxSkip-Local}}
\newcommand{\ourcomm}{\algname{Sparse-ProxSkip}}
\newcommand{\ourcommmod}{\algname{Sparse-ProxSkip-modified}}
\newcommand{\ourglobal}{\algname{Accelerated-Server-Pruning}}
\newcommand{\ourglobalmod}{\algname{Accelerated-Server-Pruning-modified}}

\usepackage{pifont}
\definecolor{mygreen1}{rgb}{0,0.8,0}
\newcommand{\green}{\color{mydarkgreen}}
\newcommand{\red}{\color{mydarkred}}
\newcommand{\xmark}{\textcolor{red}{\ding{55}}}%

\usepackage{algorithm}
\usepackage{algorithmic}

\usepackage{amsthm}
\usepackage{caption}
\usepackage{subcaption}     

\usepackage{colortbl}
\definecolor{bgcolor}{rgb}{0.93,0.99,1}
\definecolor{bgcolor2}{rgb}{0.8,1,0.8}
\definecolor{bgcolor3}{rgb}{0.50,0.90,0.50}
\definecolor{verylightgray}{rgb}{0.93, 0.93, 0.93}

\usepackage{xspace}
\usepackage{scalefnt}
\definecolor{mydarkgreen}{rgb}{0,0.45,0}
\definecolor{mydarkred}{rgb}{0.75,0,0}
\definecolor{mygreen2}{RGB}{0,120,20}
 
\newcommand{\algname}[1]{{\sf\color{mydarkred}\scalefont{0.96}{#1}}\xspace}
\usepackage{booktabs}
  
\newcommand{\norm}[1]{\left\| #1 \right\|}

\newcommand{\R}{\mathbb{R}}

\usepackage[textsize=tiny]{todonotes}
\usepackage[flushleft]{threeparttable} 

\definecolor{colabel1}{RGB}{255, 127, 14}
\definecolor{colabel2}{RGB}{44, 160, 44}
\definecolor{colabel3}{RGB}{214, 39, 40}
\definecolor{colabel4}{RGB}{148, 103, 189}

\usepackage{url}

\usepackage{hyperref}

\usepackage{fullpage}
\usepackage{natbib}

\usepackage{xspace}
 
\usepackage{booktabs}
\definecolor{mydarkblue2}{RGB}{0, 0, 139}
\newcommand{\blue}{\color{mydarkblue2}}
\newcommand{\topk}{{\blue {\rm Top}K}\xspace}

\usepackage{thmtools}
\usepackage{thm-restate}
\declaretheoremstyle[
    shaded={bgcolor=\color{HTML}{E9F6FF}}
]{shadedtheorem}

\definecolor{shadedBG}{HTML}{EFFEF0}
\declaretheoremstyle[
    shaded={bgcolor=shadedBG!50}
]{shadedlemma}

\definecolor{bggreen}{HTML}{EFFEF0}
\definecolor{bgblue}{HTML}{D5EAFA}

\title{\textbf{Sparse-ProxSkip: Accelerated Sparse-to-Sparse Training in Federated Learning}}

 \author{Georg Meinhardt$^1$ \qquad Kai Yi$^{1}$ \qquad  Laurent Condat$^{1,2}$  \qquad Peter Richt\'{a}rik$^{1,2}$\\
 \phantom{xx}
 \\
$^1$Computer Science Program, CEMSE Division,\\ King Abdullah University of Science and Technology (KAUST)\\ Thuwal, 23955-6900, Kingdom of Saudi Arabia\\
 $^2$SDAIA-KAUST Center of Excellence in Data Science and \\Artificial Intelligence 
(SDAIA-KAUST AI)
}

\date{January 31, 2025}

\begin{document}

\maketitle

\begin{abstract}
    In Federated Learning (FL), both client resource constraints and communication costs pose major problems for training large models. 
    In the centralized setting, \emph{sparse training} addresses resource constraints, while in the distributed setting, \emph{local training} addresses communication costs.    
      Recent work has shown that local training provably improves communication complexity 
    through \emph{acceleration}.
    In this work we show that in FL, naive integration of sparse training and acceleration fails, and we provide theoretical and empirical explanations of this phenomenon. 
    We introduce \ourcomm, addressing the issue and implementing the efficient technique of Straight-Through Estimator pruning 
    into sparse training.
    We demonstrate the performance of \ourcomm in extensive experiments.
\end{abstract}

\tableofcontents

\section{Introduction}
Federated learning (FL) is a distributed machine learning approach that enables multiple edge devices to collaboratively train a shared model while keeping their data local \citep{fedavg, konevcny2016federated, bonawitz2017practical}.
This paradigm addresses significant privacy concerns by avoiding the need to transfer potentially sensitive data to a central server and thus can enable access to huge datasets.
Instead, local models are trained on each client's device, and only the model updates are aggregated at the server to train a shared global model. 
However, one of the main challenges in FL is the limited computational and communication resources of edge devices~\citep{caldas2018expanding}. 

Pruning is a well-known technique in the centralized setting for reducing the computational and memory costs of model training and inference \citep{han2015learning, evci2020rigging, lee2024jaxpruner}.
There are two major directions: \emph{dense-to-sparse} or \emph{sparse-to-sparse} training~\citep{liu2023sparseland}.
\emph{Dense-to-sparse (DTS)} training starts with a dense network and proceeds by systematically removing redundant or less important parameters and reduces the model size without substantially sacrificing performance. 
\emph{Sparse-to-sparse (STS)} training starts with a sparse network and usually proceeds by sparsifying and regrowing weights but keeping the sparsity constant.
Both lead to computational savings at inference time as the final model is sparse~\citep{srinivas2017training}.
But sparse-to-sparse training also leads to substantially reduced training costs as the model is sparse throughout the whole process.
Hence, a sparse-to-sparse algorithm for FL would address the resource limitation of edge devices for efficient training and inference.
Furthermore,  \citet{lee2024jaxpruner} recently showed that the \textbf{Straight-Through Estimator (STE)} technique~\citep{ste} performs favorably in terms of final model quality in FL. 
But it requires the whole training process to be dense, including all  server and client communication.

However, a key issue during training in FL are communication costs, as for every step of the optimizer the clients have to share the model updates with the server or with each other.
\emph{Local training} has emerged as the key paradigm for efficient learning which allows the participating clients to take multiple update steps before communicating with each other.
It first appeared in the popular algorithm \fedavg and showed great empirical success in applications \citep{fedavg}.
In a recent breakthrough, \citet{proxskip} introduced \proxskip, the first algorithm to be provably more communication efficient than \fedavg by employing control variates and randomization.
In a follow-up work, \citet{randprox} were able to generalize the acceleration guarantees of \proxskip to allow for multiple proxs in an algorithm called \randprox.
In the convex setting with $l_1$ regularization, \randprox  allows to obtain a sparse model while employing acceleration, although there is no guarantee on the sparsity level. 
However, in practice, $l_1$ regularization is usually outperformed by nonconvex techniques based on the $l_0$ seminorm.

\begin{figure*}
    \centering
    \includegraphics[width=\textwidth/20*9]{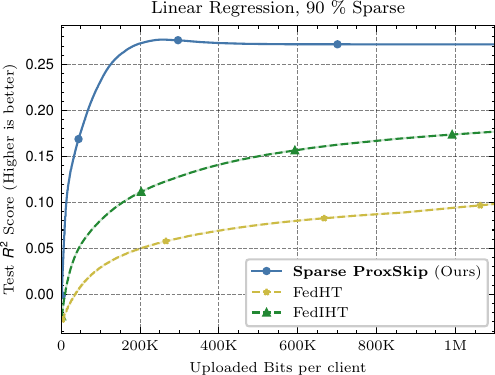}
    \includegraphics[width=\textwidth/20*9]{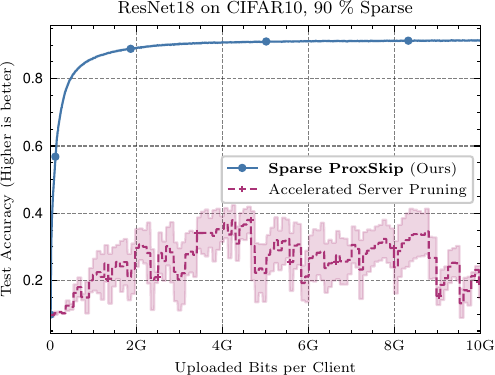}
    \caption{On the left, test score for regression on the Blog Feedback dataset~\citep{buza2013feedback}.
    Our method performs best in both final score and communication efficiency.
    On the right, test accuracy for ResNet18~\citep{he2016deep} on CIFAR-10~\citep{cifar10}.
    Our method \ourcomm prevents catastrophic failure occurring when combining
    acceleration and pruning at the server.
    The shaded area in both plots represents the standard error.
    }
    \label{fig:Convincing}
\end{figure*}

\textbf{Challenge.}\ To achieve an efficient algorithm for FL, sparse-to-sparse training and the recent theoretical advances on acceleration need to be combined. 
Hence, we address the following research question:

\textit{Is it possible to incorporate acceleration with nonconvex techniques usually found in sparse-to-sparse training algorithm?}

\textbf{Contributions.}\ A common approach in the FL literature is to apply pruning at the server~\citep{fedsparsify,lee2024jaxpruner}.
First, we show that this naive approach fails in the case of \proxskip and provide theoretical and empirical insights for this failure. 
Then, based on the theoretical guarantees by \randprox, we derive a new algorithm, \ourcomm.
Among other changes, \ourcomm combines local STE for model quality in an STS algorithm, yielding a communication efficient while powerful sparse training algorithm.
Finally, we validate our algorithm in extensive experiments.
Figure~\ref{fig:Convincing} shows how our proposed algorithm outperforms baselines for convex  and deep learning experiments.

Hence, the paper starts with an review of existing work in Section~\ref{chap:related_work}, then provides an overview of existing theoretical work on accelerated pruning in FL in Section~\ref{chap:method} and develops our method \ourcomm from that theoretical background in Section~\ref{chap:RandProxTopK} and Section~\ref{chap:proposed_method}.
Section~\ref{chap:experiments} experimentally confirms the superiority of our algorithm.
We consider linear and logistic regression  in Sections~\ref{chap:exp_regression} and \ref{chap:exp_LogR} to make comparisons with the theoretical guarantees of \randprox, since 
centralized STS regression, known as Subset Selection~\citep{hastie2017extended}, is a well established area. 
Finally, Section~\ref{chap:exp_deep_learning} deals with deep learning experiments.

\section{Related Work}
\label{chap:related_work}
Despite some existing studies on deriving sparse models in FL, the topic remains insufficiently understood.
The most similar STS approach is given by \citet{fediht}, who combine \fedavg and $\topk$ to yield \fedht and \fediht.
Their approach does not integrate acceleration or control variates.
Hence, this will be considered a baseline for our work.
Furthermore, only \fediht prunes the model before sending 
it to the server and thus uses the major communication efficiency of training a sparse model instead of a dense one~\citep{yi2024fedcomloc}.
Subsequent works do not incorporate acceleration or address client drift either~\citep{lin2022federated, bibikar2022federated, horvath2021fjord, isik2022sparse, tian2024gradient, huang2022achieving, ohib2024unmasking}, or they are not fully STS~\citep{jiang2022model, qiu2022zerofl, munir2021fedprune, li2021lotteryfl}.

In the DTS regime, the most simple approach is given by \fedsparsify, which applies Gradual Magnitude Pruning in \fedavg at the server~\citep{fedsparsify}.
The main difference between \fedht and \fedsparsify is that the latter starts with a dense model and ramps up the sparsity by a cubic schedule during the training as is usual in centralized pruning.
Another recent DTS work takes the approach of applying further centralized training approaches at the server~\citep{lee2024jaxpruner}.
Here, one gathers up the local updates (usually with a fixed learning rate) and treats them as the gradient at the server.
Then one can apply both centralized optimizers and centralized pruning techniques.
In particular, \citet{lee2024jaxpruner} apply the DTS techniques of random pruning, saliency pruning~\citep{molchanov2016pruning}, GMP~\citep{zhu2017prune} and Straight Through Estimation~\citep{bengio2013estimating} and for STS they apply static sparse training, dynamic sparse training~\citep{mocanu2018scalable} and RigL~\citep{evci2020rigging}.
We will show that acceleration and pruning at the server fail and need to be applied at the clients instead.
Hence, our work enables integrating all of the aforementioned centralized pruning techniques with \proxskip or \scaffold~\citep{scaffold}.

\section{Proposed Method}
\label{chap:method}
Our algorithm is based on the recent progress in understanding local training made in \citet{proxskip}.
Their algorithm \proxskip can optimize functions of the form
\begin{equation}
    \min _{w \in \mathbb{R}^d} f(w)+\psi(w),
    \label{eq:proxskipProblem}
\end{equation}
where $f$ is $L$-smooth and $\mu$-strongly convex and $\psi$ is proper, closed and convex \citep{bau17}. 
It corresponds to Algorithm~\ref{alg:sparse_proxskip} with the pruning options disabled. 
Under these assumptions, the optimum $w^*$ exists and is unique.
Hence, one can look at convergence against this optimum $w^*$.
Let $w^{0}$ be the initial model estimate and $w^{t}$ be the iterate of their algorithm after $t$ steps.
They proved that to be $\epsilon$ close to the optimum, i.e.~$\norm{w^{t} - w^*} \leq \epsilon \norm{w^{0} - w^*}$, one needs to evaluate the proximity operator (prox) of $\psi$ only $\sqrt{\frac L {\mu}} \log \frac 1 \epsilon$ times, while the best known bounds for Gradient Descent (and thus especially \fedavg)
is $\frac L{\mu} \log \frac 1 \epsilon$. 
One main application of \proxskip to FL  is
\[\min _{w \in \mathbb{R}^d}\left\{f(w):=\frac{1}{N} \sum_{i=1}^N f_i(w)\right\} , \]
where $f_i: \mathbb{R}^d \rightarrow \mathbb{R}$ is the loss function of each client and $N$ is the total number of clients.
This approach is closely related to empirical-risk minimization \citep{shalev2014understanding}, the dominant approach in supervised machine learning.
In practice, $f_i$ is the individual loss function of Client $i$, based on their private and local data. 
This problem is a particular case of \eqref{eq:proxskipProblem}, using a consensus formulation \citep{parikh2014proximal}.
That is, the model $w \in \R^d$ is duplicated into $N$ independent copies $w_1, w_2, \ldots, w_N$
and the objective is changed to 
\[\min _{w_1, \ldots, w_N \in \mathbb{R}^d} \frac{1}{N} \sum_{i=1}^n f_i\left(w_i\right)+\psi\left(w_1, \ldots, w_N\right),\]
where $\psi:\left(w_1, \ldots, w_N\right)\mapsto \{0$ if  $w_1=\cdots=w_N$,  $+\infty$ otherwise$\}$. 
The  proper closed convex function $\psi$ encodes the consensus constraint and the theory of \proxskip applies. 
The prox  of $\psi$ is $\operatorname{prox}_{\gamma \psi}\left(w_1, \ldots, w_N\right)=(\bar{w}, \ldots, \bar{w}) \in \mathbb{R}^{N d}$, where 
$\bar w$ is the average of the $w_i$.
Thus, evaluating the prox boils down to communicating all local models $w_1, w_2, \ldots, w_N$ to a central server and averaging them. 
Hence, one prox evaluation corresponds exactly to one communication round, the main bottleneck in   in FL~\citep{fedavg}.
Thus, reducing the number of prox evaluations is crucial to accelerate  FL, which is why \proxskip is such an important achievement for FL.

\subsection{Baseline Methods}
Additionally to \fedht and \fediht discussed in the Section~\ref{chap:related_work}, we consider 
the following simple baselines of how to address the research question of incorporating pruning, acceleration and tackling client drift. 
A  simple approach is to  employ an accelerated algorithm like \proxskip to obtain the dense solution $w^*$
and then take $\topk(w^*)$ of it for the desired sparsity, where the $\topk$ operator keeps the  $K$ largest elements of a vector unchanged and sets the other ones to zero. This approach does not address resource constraints of the clients or take advantage of training a sparse model to reduce communication cost.
We will call this approach \finaltopk.
The experiments will show that \ourmethod addresses client resources and outperforms this method, showing that it provides a valuable contribution.

Another approach would be to consider pruning at the server, i.e.\ applying $\topk$ after averaging the model
and before sending it back to the clients.
Applying optimization techniques at the server is a common approach in FL~\citep{lee2024jaxpruner, lin2022federated, fedsparsify}.
When applied to \proxskip, we refer to this variant as \ourglobal and it can be found in Algorithm~\ref{alg:sparse_proxskip}.
A major drawback is that this method does not benefit from compression for saving on uplink communication costs.
As pruning is done before downlink communication, the models uploaded to the server are dense, incurring full communication cost.
Furthermore, we show in the experiments that \ourglobal violates a key invariant of control variates, so that it is essentially inappropriate for FL.

\subsection{Accelerated Pruning Method for FL with $l_1$ regularization}
\label{chap:theory_randproxlone}
Recently, \citet{randprox} extended the framework of \proxskip to allow for several proxs while keeping acceleration.
In FL this means their algorithm \randprox can optimize problems of the form
{\small\begin{equation*}
\min _{w_1, \ldots, w_N \in \mathbb{R}^d} \frac{1}{N} \sum_{i=1}^N f_i\left(w_i\right)+\psi\left(w_1, \ldots, w_N\right) +h(w_1, \ldots, w_N),
\end{equation*}}%
for $h$ proper, closed and convex.
One interesting case is to set $h(w) = \norm{w}_1$, which comes down to \emph{federated lasso}~\citep{barik2023recovering}.
This model is known practically and theoretically to perform some sort of pruning, since  it reduces the number of nonzero parameters \citep{barik2023recovering}.
Furthermore, the $l_1$ norm is convex, so that for convex loss functions $f_i$ the accelerated convergence guarantees of  \randprox hold.
We refer to this sparse training method as \randproxlone.

\subsection{Nonconvex Modifications: Cardinality Constraints}
\label{chap:RandProxTopK}
In practice, however, it is well known that magnitude-based pruning methods outperform $l_1$ regularization, because of the bias the latter introduces. 
Cardinality constraints do not have this drawback and the algorithm can obtain the optimal solution on the subspace of the nonzero variables.
Cardinality constraints can be represented in \randprox.
One can set
\[h\left(w\right):= \begin{cases}0, & \text { if } \norm{w}_0 \leq K \\ +\infty, & \text { otherwise, }\end{cases}\]
where $\norm{w}_0$ counts the number of nonzero components of $w$.
\randprox makes calls to the prox of $h$, which is the hard-thresholding operator $\topk$~\citep{blumensath2009iterative}. 
The major caveat here is that this function $h$ is nonconvex, so that the proven acceleration guarantees of \citet{randprox} do not hold.
Empirically though, algorithms designed for the convex case have been proven powerful in the nonconvex case as well.  So, we use the  theoretical guarantees in the convex case as a strong guidance toward a powerful practical algorithm for the nonconvex case.
The resulting algorithm is \ourlocal and it can be found in Algorithm~\ref{alg:sparse_proxskip}.

A complication arises in Line~\ref{alg:line:local_version} of Algorithm~\ref{alg:sparse_proxskip} where one has to decide whether to update the control variables $h$ by the pruned $\hat w$ or unpruned weights $\tilde w$. 
We show in the following that one has to take the pruned weights, as otherwise the algorithm diverges both in theory and in practice.
To see this, first notice that the change in pseudocode is subtle.
One either takes Line~\ref{alg:line:local_version} of Algorithm~\ref{alg:sparse_proxskip} to be either
\[{\red h_{i,t+1}} = {\red h_{i,t}} + \frac{p}{\gamma}(w_{i,t+1} - \hat w_{i,t+1})\]
or 
\[ {\red h_{i,t+1}} = {\red h_{i,t}} + \frac{p}{\gamma}(w_{i,t+1} - \tilde w_{i,t+1}).\]
One can check, analogous to \citet{proxskip}, that taking the pruned weights $\hat w$ keeps the guarantee of $\sum_i h_i = 0$, while for the other choice no such guarantee holds.
We now show divergence in case of $\sum_i h_i \neq 0$.
To see this, let us look at the simple case of $p=1$ and $w_{i,0} = w^*$ for every $i$; that is, just taking one local step when being at the optimum.
Now consider the server aggregation step, i.e.~Line~\ref{alg:ProxWithControl} of Algorithm~\ref{alg:sparse_proxskip}:
\begin{align*}
    \frac 1 N \sum_{i=1}^N w^* - \gamma (g_i(w^*) - h_i) &=
w^* + \frac \gamma N \sum_{i=1}^N (g_i(w^*) - h_i) \\
&= w^* +  \frac \gamma N \sum_{i=1}^N h_i \\
&\neq w^* \quad \text{if} \quad \sum_i h_i \neq 0.
\end{align*}
The equality holds because $\sum_{i=1}^N g_i(w^*) = 0$, by first-order optimality conditions.
Hence, $w^*$ is not a fixed point and the algorithm diverges instead.
We confirmed this divergence empirically for regression and logistic regression 
and provide a detailed analysis for logistic regression in Section~\ref{chap:LogR_sum_hi}.

\subsection{Further Modifications and Proposed Algorithm}
\label{chap:proposed_method}

Furthermore, \citet{lee2024jaxpruner} recently showed the superior performance in FL of STE, compared to magnitude based pruning and in particular $\topk$.
STE approximates the Jacobian of a non-differentiable function to be the identity matrix $I$.
Hence, to apply STE to pruning, one incorporates $\topk$ into the forward pass of the model while updating the dense weights, see also \citet{bnn}.
The resulting change to the algorithm is remarkably simple, see Line~\ref{alg:line:ste} of Algorithm~\ref{alg:sparse_proxskip}.

A major problem of STE in FL is that it requires communicating dense models, hence it is not communication-efficient.
Hence, we propose to combine these two pruning methods:
For local steps on the clients, use STE  as no further communication cost is incurred.
But before communication, apply \topk to guarantee saving on communication cost.
In experiments, we noticed that this method outperforms the simple combination of \proxskip and \topk on IID data, but in deep learning on non-IID data struggles with the randomization. 
A key observation was that the algorithm performs well when the number of local steps $k$ is high $k \geq \frac 1 p$, but struggles when $k \ll \frac 1 p$.
Hence, our proposed method takes $k = \frac 1 p$ as in \fedavg or \scaffold.
The resulting algorithm is \ourcomm, found in Algorithm~\ref{alg:sparse_proxskip}.

Finally, STE is computationally expensive. 
Hence, if local computation cost is an issue we propose \ourlocal, which instead of STE applies \topk locally.
This ensures a local sparse model at the trade-off in final performance.
The resulting algorithm is \ourlocal found in Algorithm~\ref{alg:sparse_proxskip}.
For a practical application, we propose combining both methods.
Prune as little as necessary during local steps to meet local resource requirements and apply further STE for a smaller model, saving on communication cost during training and inference time compute.

We also investigated voting, saliency pruning and other pruning criteria, but found them to be non-beneficial in our experimental settings.

\begin{algorithm*}[t!]
	\caption{\algname{Meta Sparse-ProxSkip}}
    \label{alg:sparse_proxskip}
	
	\let\oldwhile\algorithmicwhile
	\renewcommand{\algorithmicwhile}{\textbf{in parallel on all workers $i \in [N]$}}
	\let\oldendwhile\algorithmicendwhile
	\renewcommand{\algorithmicendwhile}{\algorithmicend\ \textbf{local updates}}
	\begin{algorithmic}[1]
		\STATE stepsize $\gamma > 0$, probability $p>0$, initial iterate $w_{1,0}=\dots = w_{N,0}\in \R^d$, initial control variates ${\red h_{1,0}, \dots, h_{n,0}}  \in \R^d$  on each client such that  $\sum_{i=1}^{N}{\red h_{i,0}} = 0$, number of iterations $T\geq 1$
		\STATE \textbf{server:} 
        \STATE {Option \ourlocal:}  flip a coin, $\theta_t \in \{0,1\}$, $T$ times,  where $\mathop{\rm Prob}(\theta_t =1) = p$
        \STATE {Option \ourcomm:}  $\theta_i = 1$ if $ \left( i \bmod \left \lfloor \frac 1 p \right \rfloor \right) = 0 $ else $0$
	    \STATE send the sequence  $\theta_0, \dots, \theta_{T-1}$ to all workers 
		\FOR{$t=0,1,\dotsc,T-1$}
		\WHILE{}
        \STATE {Option \ourcomm}: $\tilde w_{i,t+1} = w_{i,t} - \gamma (\nabla f_i (\topk(w_{i,t})) - {\red h_{i,t}})$ \hfill $\diamond$ { \color{gray} \small STE : Prune model in forward pass only }
        \alglinelabel{alg:line:ste}
        \STATE {Option \ourlocal}: $\tilde w_{i,t+1} = \topk(w_{i,t} - \gamma (\nabla f_i (w_{i,t}) - {\red h_{i,t}}))$ 
        
		\IF{$\theta_t=1$} 
        \STATE $\hat w_{i,t+1} = \topk (\tilde w_{i, t+1})$
        \alglinelabel{alg:line:comm_mod}
		\STATE  $w_{i,t+1} = \frac{1}{N}\sum \limits_{j=1}^N \hat w_{j,t+1}$ \hfill  
        \hfill $\diamond$ { \color{gray} \small Communication with the server}
		\alglinelabel{alg:ProxWithControl}
		\STATE ${\red h_{i,t+1}} = {\red h_{i,t}} + \frac{p}{\gamma}(w_{i,t+1} - \hat w_{i,t+1})$ \hfill $\diamond$ { \color{gray} \small Update the local control variate ${\red h_{i,t}}$}\alglinelabel{alg:line:local_version}
        \alglinelabel{alg:line:global_mod_switch}
		\ELSE
		\STATE $w_{i,t+1} = \hat w_{i,t+1}$ \hfill $\diamond$ { \color{gray} \small  Skip communication!}
        \STATE $h_{i,t+1} = h_{i,t}$
		\ENDIF
		\ENDWHILE
		\ENDFOR
		\STATE $w_{i,T} = {\topk (w_{i, T})}$
		\label{ProxFinalIterate}
	\end{algorithmic}
\end{algorithm*}

\section{Experiments}
\label{chap:experiments}
We start with convex experiments 
for the following reasons.
First, the convex setting is well understood and the theoretical guarantees of \proxskip and \randprox hold only in this case.
From a theoretical point of view, $\topk$ is not nonexpansive and hence might lead to divergence. 
Hence, we start with the convex setting  to clearly investigate the effects of the mechanisms.
Second, convex models are still surprisingly widespread in industrial applications.
Third, many successful methods for the nonconvex case were designed for the convex case and then adapted to the nonconvex case. 
And lastly, \proxskip and related accelerated methods are even without pruning still underexplored in deep learning settings.
Hence, adapting these methods for sparse deep learning is challenging, but we provide experiments and general insights for this setting as well.
General experimental details can be found in Appendix~\ref{chap:general_exp}.

\subsection{Multiple Linear Regression on BlogFeedback}
\label{chap:exp_regression}

\textbf{Setup.}\ 
The first experiments tackle multiple linear regression on the BlogFeedback dataset~\citet{buza2013feedback}.
We chose this dataset for providing a realistic example of a regression problem with a natural but challenging FL split.
Previously, it has been used by \citet{barik2023recovering} to study the federated lasso, which also addresses the challenge of feature selection in a federated regression problem.
The total number of data points is $n = 47157$ split in a very heterogenous way across $554$ clients.
Furthermore, all results have been obtained by running a random search to tune the number of local steps $\frac 1 p$ and the learning rate $\gamma$.
Error bars are obtained by running the same combinations $5$ times for the same parameters with different random initialization if applicable.
More details on the dataset and the experimental setup can be found in Appendix~\ref{chap:BlogFeedbackDataset}.

\begin{figure*}[t]
	\centering
	\includegraphics[width=\textwidth/300*185]{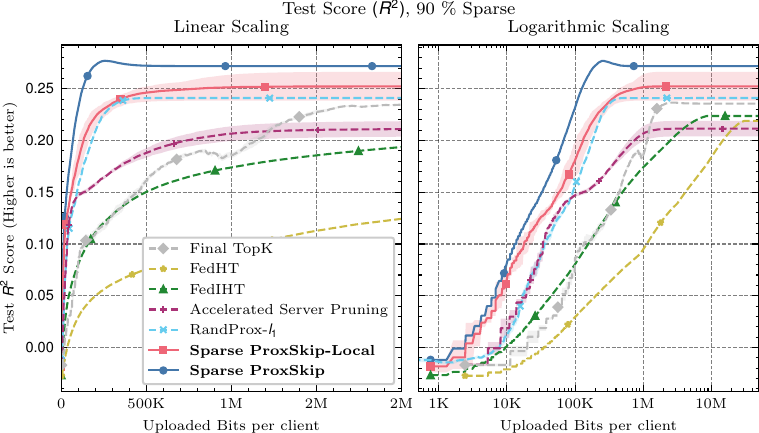}
	\includegraphics[width=\textwidth/300*97]{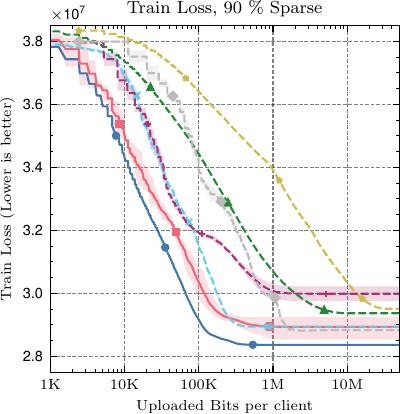}
	\caption{ Test Score ($R^2$) on the left and train loss on the right for regression on the Blog Feedback dataset~\citep{buza2013feedback}.
    Baseline methods are dashed while our methods are solid.
    We observe that both \randproxlone and our proposed methods converge to a better solution in a substantially more communication efficient way.
    The shaded area in the figures represents the standard error.
    Error bars for all experiments are included but are sometimes not visible, due to deterministic initialization at $w_{i,0} = \mathbf{0}$.
    }
	\label{fig:BlogFeedback_trajectory}
\end{figure*}

\begin{table}[t]
	   \caption{Multiple linear regression results. 
       \ourcomm shows an increase in $R^2$ due to addressing client drift.
       Table~\ref{table:BlogFeedback_acc_loss} (in the Appendix) additionally reports the final train loss.
       Results were obtained running a random search for $\gamma$ and  $p$ for all algorithms.
       }
        \label{table:BlogFeedback_acc}
        \vskip 0.15in
        \centering
        \small
		\begin{tabular}{clccc}
            \toprule
            & Sparsity & 80 \% & 90 \% & 95 \% \\
            \cmidrule(lr){3-5}
            & & Test $R^2$ & Test $R^2$ & Test $R^2$ \\
            \midrule
            \multirow{5}{*}{\rotatebox{90}{Existing}} &
            \finaltopk & $26.4$ \%& $23.8$ \% & $16.4$ \% \\
            & \fedht & \red{$18.0$} \% & \red{$21.9$} \% & \red{$12.8$} \% \\
            & \fediht & \red{$16.5$} \% & \red{$22.4$} \% & \red{$12.3$} \% \\
            &\algname{Accel.-Server-Pruning} & \red{$25.9$} \% & \red{$20.4$} \% & \red{$16.3$} \% \\
            \cmidrule(lr){2-5}
            & \randproxlone & \red{$26.3$} \% & \green{$24.1$} \% & \green{$18.8$} \% \\
            \midrule
            \multirow{2}{*}{\rotatebox{90}{Ours}} & 
            \ourlocal & \green{$27.0$} \% & \green{$26.8$} \% & \green{$23.9$} \% \\
            & \ourcomm & \green{$27.7$} \% & \green{$27.7$} \% & \green{$26.5$} \% \\
            \bottomrule
        \end{tabular}
\end{table}

\begin{table*}[t]
	   \caption{Communication cost to reach a certain test $R^2$ score for multiple linear regression at $90 \%$ sparsity.
        All speedup comparisons are with respect to \finaltopk as it is an accelerated method outperforming \fediht and is the only baseline reaching a test score of $0.225$.
        }
        \label{table:BlogFeedback_comm}
             \vskip 0.15in
        \centering
        \small
		\begin{tabular}{clcccccc}
            \toprule
			& Test $R^2$ Threshold & \multicolumn{2}{c}{$0.2$} & \multicolumn{2}{c}{$0.225$} & \multicolumn{2}{c}{$0.25$} \\
            \cmidrule(lr){3-4} \cmidrule(lr){5-6} \cmidrule(lr){7-8}
			& Upload Communication Cost & Bits & Speedup & Bits & Speedup & Bits & Speedup \\
			\midrule
			\multirow{5}{*}{\rotatebox{90}{Existing}} 
            & \finaltopk & $1.16$ M & $1.00\times$ & 1.44 M & 1.00$\times$ & \xmark & \xmark \\
			& \fedht & $14.8$ M & {\red 0.08}$\times$ & \xmark & \xmark & \xmark & \xmark \\
			& \fediht & $2.49$ M & {\red 0.47}$\times$ & \xmark & \xmark & \xmark & \xmark \\
			& \ourglobal & $0.73$ M & {\green 1.59}$\times$ & \xmark & \xmark & \xmark & \xmark \\
            \cmidrule(lr){2-8}
			& \randproxlone & $0.18$ M & {\green 6.44}$\times$ & 0.25 M  & {\green 5.76}$\times$ & \xmark &  \xmark \\
            \midrule
			\multirow{2}{*}{\rotatebox{90}{Ours}}
            & \ourlocal & $0.13$ M  & {\green {8.90}}$\times$ & 0.21 M  & {\green 6.86}$\times$ & 0.76 M & - \\
			& \ourcomm & $0.07$ M &  {\green 16.6}$\times$ & 0.09 M & {\green 16.0}$\times$ & 0.13 M & - \\
            \bottomrule
		\end{tabular}
\end{table*}

\textbf{Experimental Results.}\ 
Our methods improves both in $R^2$ (quality of the solution) and in communication efficiency over the baselines.
Training trajectories for a sparsity of $90 \%$, showing the gains in communication cost and accuracy at the same time,  can be found in Figure~\ref{fig:BlogFeedback_trajectory}.
Table~\ref{table:BlogFeedback_acc} reports the final $R^2$ (solution quality) for different target sparsity values.
At $90 \%$ sparsity, we see that \ourcomm improves by $3.9 \%$ over the best baseline \finaltopk and $5.3 \%$ over the best non-client-drift-addressing variant.
Furthermore, the advantage grows with increased sparsity at $95 \%$.
Table~\ref{table:BlogFeedback_comm} reports the gains in communication efficiency.
We can observe that \ourcomm is roughly $16\times$ more communication efficient than the best baseline \finaltopk and roughly $32\times$ more communication efficient than the 
best non-accelerated baseline.

\textbf{RandProx-$l_1$ Beats Simple Baselines.}\ 
We see that \randproxlone, as described in Section~\ref{chap:theory_randproxlone}, outperforms the simple baselines in terms of both communication efficiency and $R^2$.

Noticeably, this supports our hypothesis in that: 
1) Acceleration (through \randproxlone) leads to a communication cost decrease of $\geq 6 \times$ compared to \fediht.
2) Addressing client drift (through \randproxlone) leads to an increase in final test score of up to $2.4 \%$ compared to \fediht.
3) \randproxlone outperforms naive baselines like pruning at the server or pruning at the end, showing the need for a properly designed accelerated STS method. 

\textbf{Failure of \emph{Accelerated Server Pruning}.}\ 
From Figure~\ref{fig:BlogFeedback_trajectory}, we observe that \ourglobal performs worst from all tested baselines.
In particular, it performs worse than \fediht which does neither address client drift nor is accelerated.
As we discussed in Section~\ref{chap:proposed_method} we hypothesized this because the property $\sum_i h_i \neq 0$ is violated in \ourglobal.
We confirmed this hypothesis empirically for logistic regression 
and provide a detailed analysis in Section~\ref{chap:LogR_sum_hi}.

\textbf{Sparse-ProxSkip-Local beats RandProx-$l_1$.}\ 
We finally note that \ourcomm outperforms \randproxlone and the other baselines.
We make the following observations:
1) \randproxlone reaches the desired sparsity only gradually.
The theory only guarantees convergence to a sparse solution, but there is no guarantee during the training.
Hence, the communication costs it occurs are larger than when applying $\topk$ for local pruning.
2) One can notice an accuracy improvement of \ourlocal compared to \randprox $l_1$.
We attribute this to the bias induced by $l_1$ regularization.

\subsection{Multiple Logistic Regression on FEMNIST}
\label{chap:exp_LogR}

\begin{figure*}[t]
	\centering
	\includegraphics[width=\textwidth/300*185]{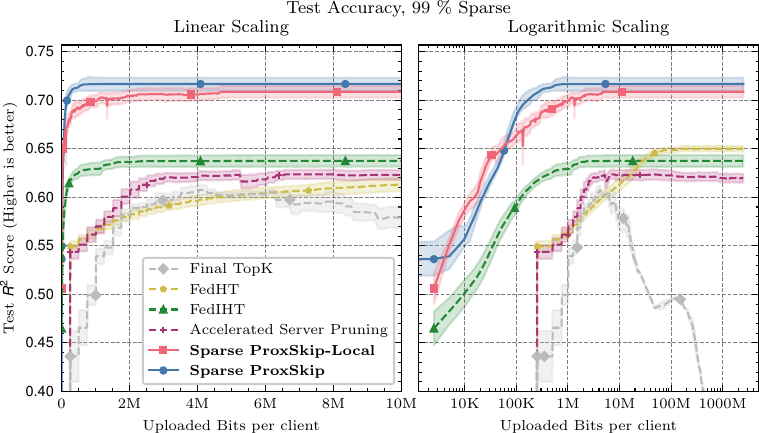}
	\includegraphics[width=\textwidth/300*97]{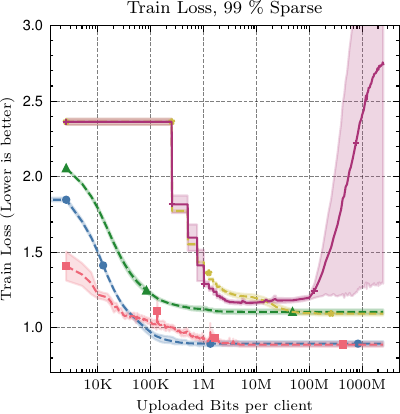}
	\caption{Results for logistic regression on FEMNIST at $99 \%$ sparsity.
        \ourcomm and \ourlocal outperform all baselines both in communication costs and final accuracy.
        The shaded area in the figures represents the standard error.}
	\label{fig:LogR_results}
\end{figure*}

\begin{table*}[t]
    \caption{Communication costs to reach a certain test accuracy at $90 \%$ sparsity on FEMNIST.        Note that although the final accuracy for \ourglobal is below 80\% as seen in Table~\ref{table:LogR_acc}, it peaks at 84 \% early on.
        The same holds for \finaltopk and 85 \%.
    }
    \label{table:LogR_comm}
      \vskip 0.15in
        \centering
        \small
        \begin{tabular}{clcccccc}
            \toprule
			& Test Accuracy Threshold & \multicolumn{2}{c}{80 \%} & \multicolumn{2}{c}{82.5 \%} & \multicolumn{2}{c}{85 \%} \\
            \cmidrule(lr){3-4} \cmidrule(lr){5-6} \cmidrule(lr){7-8}
			& Upload Communication Cost & Bits & Speedup & Bits & Speedup & Bits & Speedup \\
			\midrule
			\multirow{5}{*}{\rotatebox{90}{Existing}} 
            & \finaltopk & 6.0 M & {\red 0.1}$\times$ & 16.6 M & {\red 0.1}$\times$  & 92.4 M & {\red 0.1}$\times$ \\
			& \fedht & 4.0 M & {\red 0.2}$\times$ & 8.54 M & {\red 0.2}$\times$ & 45.7 M & {\red 0.3}$\times$  \\
			& \fediht & 0.8 M & 1.0$\times$ & 1.61 M & 1.0$\times$ & 13.0 M & 1.0$\times$ \\
			& \ourglobal & 2.0 M & {\red 0.4}$\times$ & 3.57 M & {\red 0.5}$\times$  & \xmark & \xmark \\
            \midrule
			\multirow{2}{*}{\rotatebox{90}{Ours}}
            & \ourlocal & 0.5 M & {\green 1.8}$\times$ & 0.55 M & {\green 2.9}$\times$ & 2.52 M & {\green 5.2}$\times$ \\
			& \ourcomm & 0.1 M & {\green 8.0}$\times$ & 0.18 M & {\green 8.9}$\times$ &  0.36 M & {\green 36}$\times$ \\
            \bottomrule
		\end{tabular}
\end{table*}

\textbf{Setup.}\ 
A more challenging but still convex setting is multiple logistic regression on the FEMNIST dataset~\citep{caldas2018leaf}.
We take the naturally-occurring federated split but limit the number of clients to $N =100$.
A similar approach was taken by \citet{jiang2022model} for $N = 193$.
The reasoning and further details can be found in Appendix~\ref{chap:DetailsLogR}.

\textbf{Results.}\ 
The general results are shown in Figure~\ref{fig:LogR_results}.
Results on communication efficiency are reported in Table~\ref{table:LogR_comm}.
As only \fediht enjoys communication speedup from compression, it is taken as the baseline so that the reported speedup is solely due to acceleration.
We see that \ourlocal is $1.2$--$5.2\times$ more communication efficient and \ourglobal is $4$--$20\times$ more communication efficient than \fediht.
If \fedht is taken as the baseline, which would be a usual approach for obtaining pruned models in FL~\citep{lee2024jaxpruner}, then \ourlocal is $9$--$18\times$ and \ourglobal is $20$--$70\times$ more communication efficient than \fedht.

Results on the final accuracy for different sparsity levels are reported in Table~\ref{table:LogR_acc}.
We observe that the advantage of our method is significant only with high sparsity levels. 
That is,~at 80 \% there is just a $0.3 \%$ advantage, while at 99 \% the gap has widened to 5.3 \%.
On the other hand, for sparsity $80$ \% and $90$ \% the performance of \finaltopk is competitive with 
the other methods. 
This suggests that achieving these sparsity levels is not challenging on FEMNIST.

\begin{table*}[t]
	   \caption{Test accuracy of logistic regression on FEMNIST for different sparsity levels.
            The best accuracy for each sparsity level is highlighted in bold.
        }
        \label{table:LogR_acc}
             \vskip 0.15in
        \centering
        \small
		\begin{tabular}{clccccc}
            \toprule
			& Sparsity & 80 \%& 90 \% & 95 \% & 98 \% & 99 \%\\
            \midrule
			\multirow{4}{*}{\rotatebox{90}{Existing}} & 
                \finaltopk & \red{84.7} \% & \red{79.9} \% & \red{69.6} \% & \red{40.1} \% & \red{25.5} \%  \\
			& \fedht & 86.6 \% & 85.7 \% & 84.7 \% & 76.6 \% &  66.4 \% \\
			& \fediht & \green{86.8} \% & \red{85.6} \% & \red{82.7} \% & \red{74.6} \% &  \red{65.4} \% \\
			& \ourglobal & \red{77.9} \% & \red{77.5} \% & \red{76.8} \%  & \red{72.2} \% & \red{64.7} \%  \\
            \midrule
			\multirow{2}{*}{\rotatebox{90}{Ours}} & 
            \ourlocal & \green{86.7} \% & \green{86.1} \% & \green{84.7} \% & \green{78.9} \% &  \green{70.7} \% \\
			& \ourcomm & \green{\textbf{86.9}} \% & \green{\textbf{86.7}} \% & \green{\textbf{85.0}} \% & \green{\textbf{79.3}} \% &  \green{\textbf{71.7}} \% \\
            \bottomrule
		\end{tabular}
\end{table*}

\begin{figure*}[t]
	\centering
	\includegraphics[width=\textwidth/300*180]{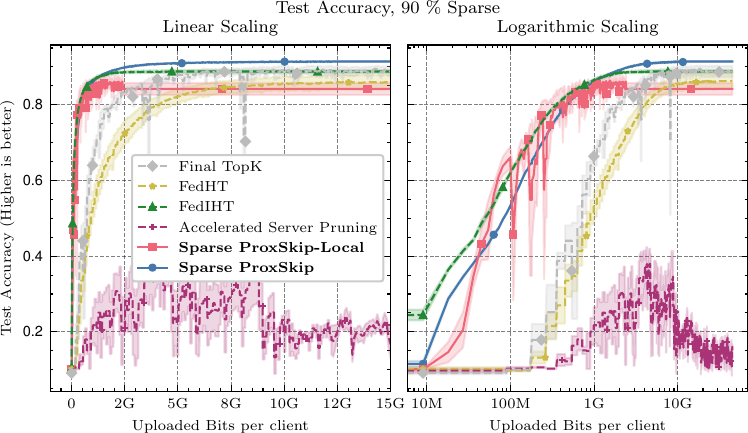}
	\includegraphics[width=\textwidth/300*98]{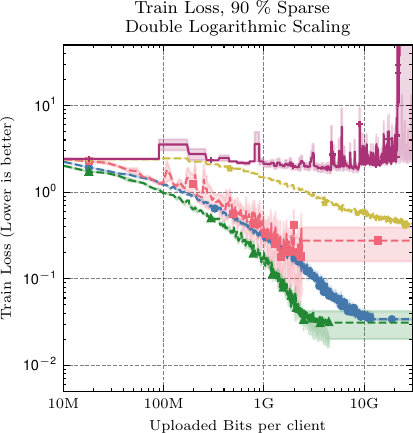}
    \caption{Results for ResNet18~\citep{he2016deep} on CIFAR10~\citep{cifar10} at $90 \%$ sparsity.
        \ourcomm  still outperforms the baselines,  to a lesser degree though.
        The main observation is that \ourglobal fails completely in accuracy and loss because of $|\sum_i h_i| \gg 0$ and that the proposed fixes of \ourcomm address this problem.
        The shaded area in the figures represents the standard error.
        }
	\label{fig:dnn_trajectory}
\end{figure*}

\subsubsection{Pruning and control variables}
\label{chap:LogR_sum_hi}
In Section~\ref{chap:RandProxTopK} we proved that one has to prune communicating / updating the control variables, as otherwise the algorithm might diverge. 
The crucial observation made in Section~\ref{chap:RandProxTopK} is that if $\sum_i h_i \neq 0$, then the algorithm diverges.
We experimentally confirmed on logistic regression for FEMNIST that $\sum_i h_i \neq 0$  leads to impaired performance on real world datasets and $|\sum_i h_i| \gg 0$ holds for \ourglobal.
Details are found in Appendix~\ref{chap:zero_sum_control}.
This also shows that any pruning at the server combined with control variables will fail; for instance, one should expect similar results when combining \scaffold with \topk at the server.

\subsection{Deep Learning Experiments}
\label{chap:exp_deep_learning}

Further nonconvex experiments were conducted on CIFAR10~\citep{cifar10} using ResNet18~\citep{he2016deep}.
Further details can be found in Appendix~\ref{chap:appendix_dnn}.

The results for $90 \%$ sparsity are shown in Figure~\ref{fig:dnn_trajectory}.
Mainly, we note that \ourglobal fails completely both in accuracy and in the loss increasing instead of decreasing.
The algorithm does not head towards a minimum of the loss.
This is because early on, the sum of the control variates $\sum_i h_i$ grows quickly and shifts all subsequent local gradients.
Hence, one can see that keeping $\sum_i h_i = 0$ is particularly important for large models.
Furthermore, one can see that the proposed variant \ourcomm performs best and gives the highest final accuracy.
We attribute this superiority to the control variates counteracting client drift.
On the other hand, in this scenario there seems to be no benefit from acceleration.
This aligns with earlier observations that acceleration faces challenges in deep learning~\citep{defazio2019ineffectiveness} and that addressing client drift proves beneficial for final accuracy nonetheless~\citep{li2023effectiveness}.
However, \citet{li2023effectiveness} found that control variates also benefit to the communication cost in highly heterogenous settings.
While we applied the same federation process as \citet{li2023effectiveness}, our different observations might be due to the different levels of participation and number of clients. Indeed a different amount of data per client induces a different level of heterogeneity for the Dirichlet distribution with parameter $\alpha$.

\section{Conclusion}
\label{chap:conclusion}

We investigated whether it is possible in FL to combine acceleration with sparse training.
We showed that the naive combination of these techniques fails and that it is theoretically and empirically crucial to keep the sum of the control variates, that correct client drift, to zero.
Based on these important findings, we developed a theoretically-motivated method, \ourcomm, which integrates the successful mechanism of $\topk$ and STE for sparse training in FL. 
Furthermore, we proposed the first method to integrate STE, which is prohibitively costly in terms of communication, into a communication-efficient STS training method.
Our experiments confirm the efficiency of our proposed \ourcomm method.

\section*{Acknowledgments} This work was supported by funding from King Abdullah University of Science and Technology (KAUST): i) KAUST Baseline Research Scheme, ii) Center of Excellence for Generative AI, under award number 5940, iii) SDAIA-KAUST Center of Excellence in Data Science and Artificial Intelligence.

\bibliographystyle{plainnat}
\bibliography{egbib}

\newpage
\appendix
\onecolumn

 \noindent{\huge \textbf{Appendix}}

\begin{table*}[h]
	   \caption{Blog Feedback Dataset results. Results were tuned for $\gamma$ and  $p$ and hence show the improved scores due to addressing client drift.
           }
        \label{table:BlogFeedback_acc_loss}
        \vskip 0.15in
        \centering
        \small
        \setlength{\tabcolsep}{3pt}
		\begin{tabular}{clcccccc}
            \toprule
            & Sparsity & \multicolumn{2}{c}{80 \%} & \multicolumn{2}{c}{90 \%} & \multicolumn{2}{c}{95 \%} \\
            \cmidrule(lr){3-4} \cmidrule(lr){5-6} \cmidrule(lr){7-8}
            & & Train Loss & Test $R^2$ & Train Loss & Test $R^2$ & Train Loss & Test $R^2$ \\
            \midrule
            \multirow{5}{*}{\rotatebox{90}{Existing}} &
            \finaltopk & $2.817e7$ & $26.4\%$ & $2.877e7$ & $23.8\%$ &  $3.113e7$ & $16.4\%$ \\
            & \fedht & $3.056e7$ & \red{$18.0$} \% &  $2.951e7$ & \red{$21.9$} \% & $3.288e7$ & \red{$12.8$} \% \\
            & \fediht & $3.143e7$ & \red{$16.5$} \% & $2.937e7$ & \red{$22.4$} \% & $3.267e7$ & \red{$12.3$} \% \\
            & \ourglobal & $2.872e7$ & \red{$25.9$} \% & $2.991e7$ & \red{$20.4$} \% & $3.217e7$ & \red{$16.3$} \% \\
            \cmidrule(lr){2-8}
            & \randproxlone & $2.823e7$ & \red{$26.3$} \% & $2.894e7$ & \green{$24.1$} \% & $3.073e7$ & \green{$18.8$} \% \\
            \midrule
            \multirow{2}{*}{\rotatebox{90}{Ours}} & 
            \ourlocal & $2.818e7$ & \green{$27.0$} \% & $2.856e7$ & \green{$26.8$} \% & $2.938e7$ & \green{$23.9$} \% \\
            & \ourcomm & $2.810e7$ & \green{$26.7$} \% & $2.831e7$ & \green{$27.1$} \% & $2.897e7$ & \green{$26.7$} \% \\
            \bottomrule
        \end{tabular}
\end{table*}

\section{General Experimental Details}
\label{chap:general_exp}
Our experiments were implemented in Python using Pytorch. 
The experiments were conducted on our local workstations equipped with Intel(R) Xeon(R) Gold 6226R CPUs (2.90 GHz), 1 TB of RAM, and four Nvidia A100 GPUs, each with 40 GB of VRAM, although much less is required to reproduce these results.
Each single training run of the experiments took no more than 20 hours of compute time.
Some methods do not produce models at the desired sparsity, e.g.~\fediht usually yields a model of $70 -90 \%$ when given a target sparsity of $90 \%$.
Hence, before any evaluation of any method the models are pruned to the target sparsity by applying $\topk$.

\section{Experimental Details: Linear Regression}
\label{chap:BlogFeedbackDataset}
\textbf{Blog Feedback Dataset Details.}\ 
The dataset contains a number of blog posts with their respective number of comments so far and the goal is to predict the number of comments over the following $24$h time window. 
For federating the dataset, it has a natural split by considering the source page 
where a particular blog post appeared, i.e.~the website domain where it was published.
For each domain, we create one client.\footnote{
    In practice this means grouping by the first 50 columns as these are attributes of the source website and creating a client for each unique combination of values in these columns}
Furthermore, before federating we scale all attributes to be in the range $[0,1]$ to make the computations more amenable.
This results in a dataset with $554$ clients.
A histogram of the client size can be found in Figure~\ref{fig:blogFeedbackHistogram} in the appendix.
To add a bias term, which is usual for regression, we modify every sample to have an additional entry $1$.

\textbf{Objective Function.}\ 
We optimize the objective function
\[f(w) = \frac 1 N \sum_{i=1} ^N f_i(w) = \frac 1 N \sum_{i=1}^N \left(\frac 1 2 \norm{A_i w - b_i}^2_2 
+ \frac \alpha {4} \norm{w}^2_2 \right) + \frac 1 {2N} \phi(w).\]
Here $\phi$ encodes our sparsity constraint, i.e.\ either $\norm{\cdot}_1$ or cardinality constraints resulting in $\topk(\cdot)$ and 
$A_i$ is the local data matrix.
$\alpha = 10^{3}$ in our experiments and was empirically chosen to give good $R^2$ on a validation set.

\textbf{Evaluation Metrics.}\ In addition to reporting the loss, the BlogFeedback dataset~\citet{buza2013feedback} contains a train and a test split.
The test split is \emph{out-of-distribution} which in this case means that the test data was recorded at least $1$ month up to a year later compared to the training dataset.
To measure the error for regression it is usual to report the $R^2$ metric which lies between $0$ and $1$
for favorable predictors.
A $R^2$ value of $0$ does not explain the dataset at all while a values of $1$ would explain the dataset fully.
Hence, a higher $R^2$ is better.

\textbf{Initialization.}\ 
Regression is a convex scenario, so that for \randprox convergence is guaranteed from any starting point. 
Thus, to induce sparsity from the beginning, the initial model is chosen as $ w_{i,0} = \mathbf{0}$ for every $i$.

\textbf{Hyperparameters.}\ 
The hyperparameters, which are the learning rate $\gamma$ and average number of local steps $\frac 1 p$ were tuned by a random search. 
First a suitable range for these parameters was identified, then in a second random search the best parameters in this range were taken for the final experiments.
Then, the average of 5 runs was taken to obtain the presented results.
All algorithms were run for $10^4$ communication rounds ensuring convergence to their respective solutions.

\textbf{Full Experimental Results.}\ 
The results for the sparsity comparison including the loss function can be found in Table~\ref{table:BlogFeedback_acc_loss}.
From the loss one can see that the optimizer is not only better at increasing $R^2$, but also at decreasing the objective function.

\begin{figure*}[t]
    \centering
    \includegraphics[width=\textwidth/2]{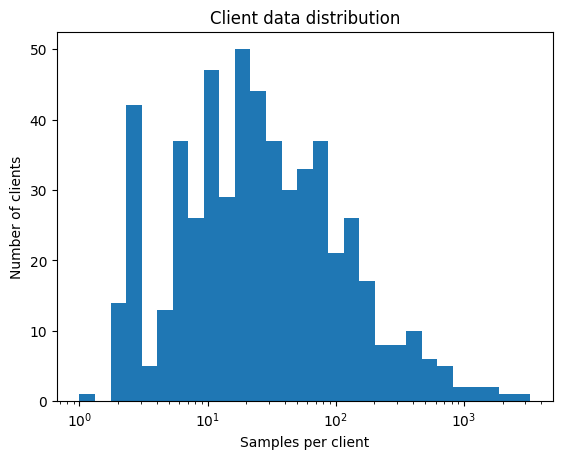}
    \caption{Distribution of the client sizes in the Federated version of the Blog Feedback dataset~\citep{buza2013feedback}.}
    \label{fig:blogFeedbackHistogram}
\end{figure*}

\section{Experimental Details: Logistic Regression}
\label{chap:DetailsLogR}
\textbf{Dataset.}\ We run the experiments on the FEMNIST dataset~\citep{caldas2018leaf}, a common benchmark of the FL community that possesses a natural federated partition.
We only consider $N = 100$ clients out of the $3220$ naturally occurring in FEMNIST for the following reasons.
A similar approach was taken by \citet{jiang2022model} for $N = 193$.
On the one hand, \proxskip requires modifications to support partial client participation~\citep{condat2023tamuna, grudzien2023can},  but in the setup chosen here only allows for full client participation. 
A high number of clients participating in each round is unrealistic~\citep{charles2021large}.
The goal of this work is to benchmark the advantage of control variates for client drift, hence providing a benchmark on natural federated splits is crucial.
Merging clients would diminish the advantage of having a realistic federated split.

On the other hand, too few clients result in too little data.
Hence, $100$ was chosen as a tradeoff between these aspects resulting in a dataset of $n = 11152$ images.
We employed the standard unrestricted test dataset.
The performance tradeoff for this choice is that our centralized dense estimator achieves an accuracy of $89.4 \%$ when trained on the full FEMNIST dataset, compared to $85.4 \%$ when trained on our restricted dataset.

\textbf{Objective Function.}\ 
We align our objective function with the one from \texttt{scikit-learn} which uses the softmax formulation; that is,
we define 
\[\hat{p}_k(\mathbf{x}_i) = \frac{\exp(\mathbf{x}_i w_k + w_{0, k})}{\sum_{l=0}^{K-1} \exp(\mathbf{x}_i w_l + w_{0, l})}\]
and minimize
\[\min_w f(w) = \frac 1 N \sum_{i=1} ^N f_i(w) = \frac 1 N \sum_{i=1} ^N \left( - \frac N n \sum_{i=1}^{n_i} \sum_{k=0}^{K-1} [y_i = k] \log(\hat{p}_k(\mathbf{x}_i)) + \frac \alpha {2} \norm{w}^2_2 \right) + \frac 1 {2N} \phi(\mathbf{w}).\]
$N$ is the number of clients, $n$ is the total number of samples and $n_i$ is the number of samples of Client $i$.
Furthermore, $\mathbf{x_i}$ refers to a single datapoint and $y_i$ is its label.

\textbf{Hyperparameters.}\ 
The hyperparameters of the learning rate $\gamma$ and local steps $\frac 1 p$ were tuned by a random search. 
First a suitable range for these parameters was identified, then in a second random search the best parameters in this range were taken for the final experiments.
Then, the average of 5 runs was taken to obtain these results.
The default initialization for a linear layer of Pytorch was taken.

\section{Zero-Sum of the Control Variates}
\label{chap:zero_sum_control}

\begin{figure*}[t]
    \centering
    \includegraphics[width=\textwidth/2]{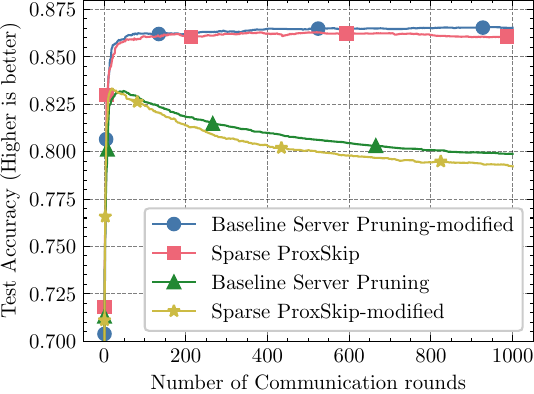}
    \caption{Test accuracy of our method and server pruning.
    The modified variants keep $\sum_i h_i = 0$.
    We can clearly see that this improves accuracy.}
    \label{fig:control_sum_test}
\end{figure*}

This section provides empirical insights on why the property $\norm{\sum_i h_i} = 0$ is crucial and its violation in \ourglobal on logistic regression with FEMNIST and $90 \%$ sparsity.
This refers to the setting and reasoning of Section~\ref{chap:LogR_sum_hi}.

First, Figure~\ref{fig:control_sum_test} shows the observation that \ourcomm outperforms \ourglobal.
As a first step we introduce the following modified variants of these two algorithms.
\ourcommmod changes Line~\ref{alg:line:global_mod_switch} of Algorithm~\ref{alg:sparse_proxskip} to be 

\[{h_{i,t+1}} = { h_{i,t}} + \frac{p}{\gamma}(w_{i,t+1} - {\red\tilde w_{i,t+1}})\] 
instead of  
\[{h_{i,t+1}} = {h_{i,t}} + \frac{p}{\gamma}(w_{i,t+1} - {\red\hat w_{i,t+1}}).\]
Or more intuitively: It uses the unpruned variables for updating the control variables instead of the pruned ones.
This has the effect of violating $\sum_i h_i = 0$.
Furthermore, \ourglobalmod now prunes on the client before any local updates but after the control variates have been updated.
This variant has no practical purpose as it does not save either on uplink or downlink communication but crucially it 
guarantees $\sum_i h_i = 0$.
Figure~\ref{fig:control_sum_test} shows that the latter is a competitive variant and fixes the issue with \ourglobal.

\begin{figure*}[t]
    \centering
    \includegraphics[width=9.5\textwidth/20]{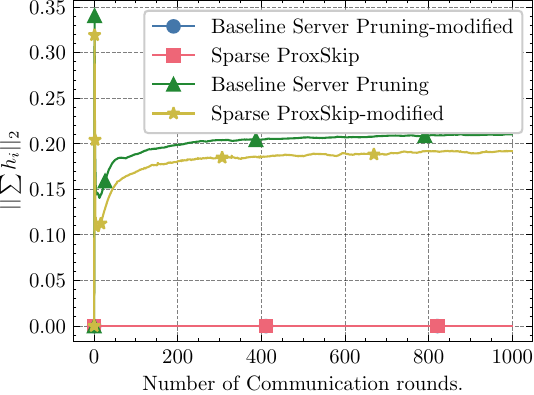}
    \includegraphics[width=9.5\textwidth/20]{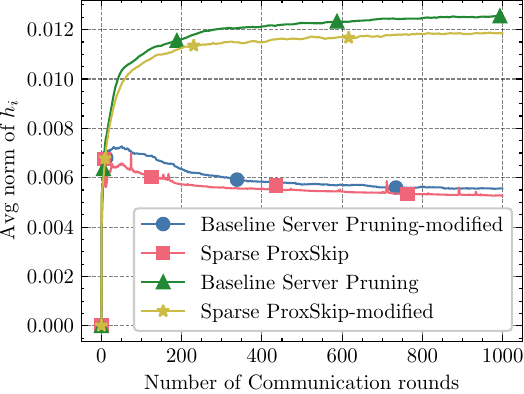}
    \caption{Norm of $\sum_i h_i$ on the left vs average norm of $h_i$ on the right.}
    \label{fig:control_sum_sum}
\end{figure*}

\begin{figure*}[t]
    \centering
    $\!\!\!$\includegraphics[width=9.5\textwidth/20]{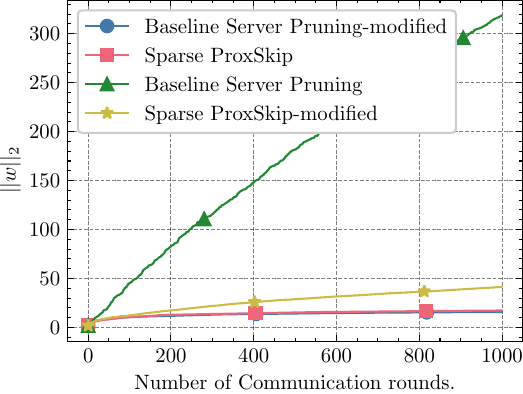}
    \includegraphics[width=9.5\textwidth/20]{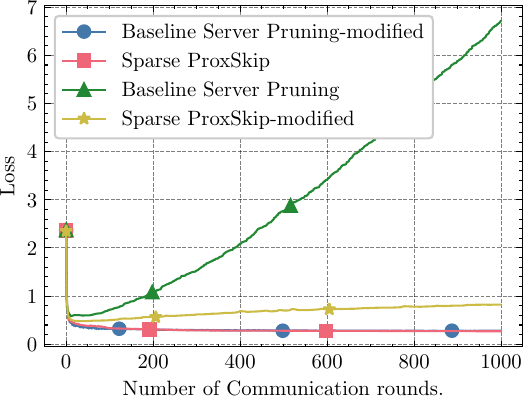}
    \caption{Norm of the model $w$ and loss value.}
    \label{fig:control_sum_params}
\end{figure*}

First, on the left in Figure~\ref{fig:control_sum_sum} one can see that $\sum_i h_i$ is far from $0$, and combined with the plot on the right on the average norm of $h_i$, one can draw the conclusion that the size of $\sum_i h_i$ dominates the control variables themselves.
Hence, with the proof from Section~\ref{chap:proposed_method} one can conclude that the algorithm diverges by shifting the gradient by $\sum_i h_i$.
To see this empirically, one can look at the norm of the parameters in Figure~\ref{fig:control_sum_params}.
Both \ourcomm and \ourglobalmod converge to roughly the same parameters norm.
The other variants though, for which $\sum_i h_i \neq 0$ holds, seem to move far away from this parameter combination.
The plot on the left in Figure~\ref{fig:control_sum_params} confirms this in the loss: instead of minimizing the loss, the methods diverge significantly.

\section{Experimental Details: Deep Learning on CIFAR10}
\label{chap:appendix_dnn}
\textbf{Experimental Details.}\ 
The experiments were run on CIFAR10~\citep{cifar10} using ResNet18~\citep{he2016deep}.
The number of clients was $N = 10$ with full client participation.
The data was distributed through a Dirichlet distribution with parameter $\alpha = 0.3$.
The number of samples per client is distributed according to a lognormal distribution with variance $0.3$.
We used \emph{FedLab} for producing the federated data split~\citep{smile2021fedlab}.
A random search was conducted to find the best parameters among learning rate, local steps, batch size and gradient clipping value.
The experiments were run for $500$ rounds for.
The number of local steps was chosen from the range $\{8, 16, 32, 64, 128, 256\}$.
For \proxskip, $p = \frac 1 {\text{\#local steps}}$ is taken.
The batch size was chosen from the range $\{32, 64 \}$.
The gradients were clipped by a value chosen log-uniformly between $10$ and $200$.
Without gradient clipping, \proxskip would run into \texttt{NaN} errors.
We used a weight decay of $10^{-4}$ and applied common transforms on the training data of flipping, cropping and normalizing.
  
\section{Outlook and Limitations}
\label{chap:limitations}
Sparse training might prove crucial for training large models in FL, which offer architectural benefits over small models.
Here, sparse training enables larger models to respect the resource requirements of edge devices.
Furthermore, these findings might be invaluable for combining centralized sparse training and pruning methods with acceleration.
We provided a general invariant that pruning has to take place at the clients but future work might address the details of this integration.
Additionally, in its current form, the method provides inference benefits and communication cost savings but would need further development for reducing the computational costs during training.
In particular, our current gradients and control variables are dense, requiring further modification before yielding a sparse-to-sparse training method with the computational and memory footprint of a small model.
In the pruning literature, masking is usually employed for this aspect.
Here, one could apply masking to the control variates as well and combine gradient calculation and pruning so as to decrease the memory cost of the full gradients.

\end{document}